\documentclass[runningheads]{llncs}
\usepackage{graphicx}
\usepackage{color}
\usepackage{marvosym}
\usepackage{multirow}
\usepackage{amssymb}
\usepackage{bbm}
\def\ie{\emph{i.e.}}
\def\eg{\emph{e.g.}}
\def\etal{\emph{et al.}}

\newcommand{\re}[1]{{\textbf{ \color{blue} #1 }}}

\begin{document}
\title{Text-Aware Single Image Specular Highlight Removal}

%
\author{Shiyu Hou\inst{1,2} 
Chaoqun Wang\inst{1,2} Weize Quan\inst{2,1} Jingen Jiang\inst{1,2} and Dong-Ming Yan\inst{2,1}\textsuperscript{\Letter}}

\authorrunning{S.Hou et al.}
%
\institute{University of Chinese Academy of Sciences, Beijing 100049, China
\and Institute of Automation, Chinese Academy of Sciences, Beijing 100190, China
\email{weize.quan@nlpr.ia.ac.cn, yandongming@gmail.com}}
\maketitle              
\begin{abstract}
Removing undesirable specular highlight from a single input image is of crucial importance to many computer vision and graphics tasks. Existing methods typically remove specular highlight for medical images and specific-object images, however, they cannot handle the images with text. In addition, the impact of specular highlight on text recognition is rarely studied by text detection and recognition community. Therefore, in this paper, we first raise and study the text-aware single image specular highlight removal problem. The core goal is to improve the accuracy of text detection and recognition by removing the highlight from text images. To tackle this challenging problem, we first collect three high-quality datasets with fine-grained annotations, which will be appropriately released to facilitate the relevant research. Then, we design a novel two-stage network, which contains a highlight detection network and a highlight removal network. The output of highlight detection network provides additional information about highlight regions to guide the subsequent highlight removal network. Moreover, we suggest a measurement set including the end-to-end text detection and recognition evaluation and auxiliary visual quality evaluation. Extensive experiments on our collected datasets demonstrate the superior performance of the proposed method.

\end{abstract}
%
%
\section{Introduction}
\begin{figure}[t]
\centering
\includegraphics[width=0.95\linewidth]{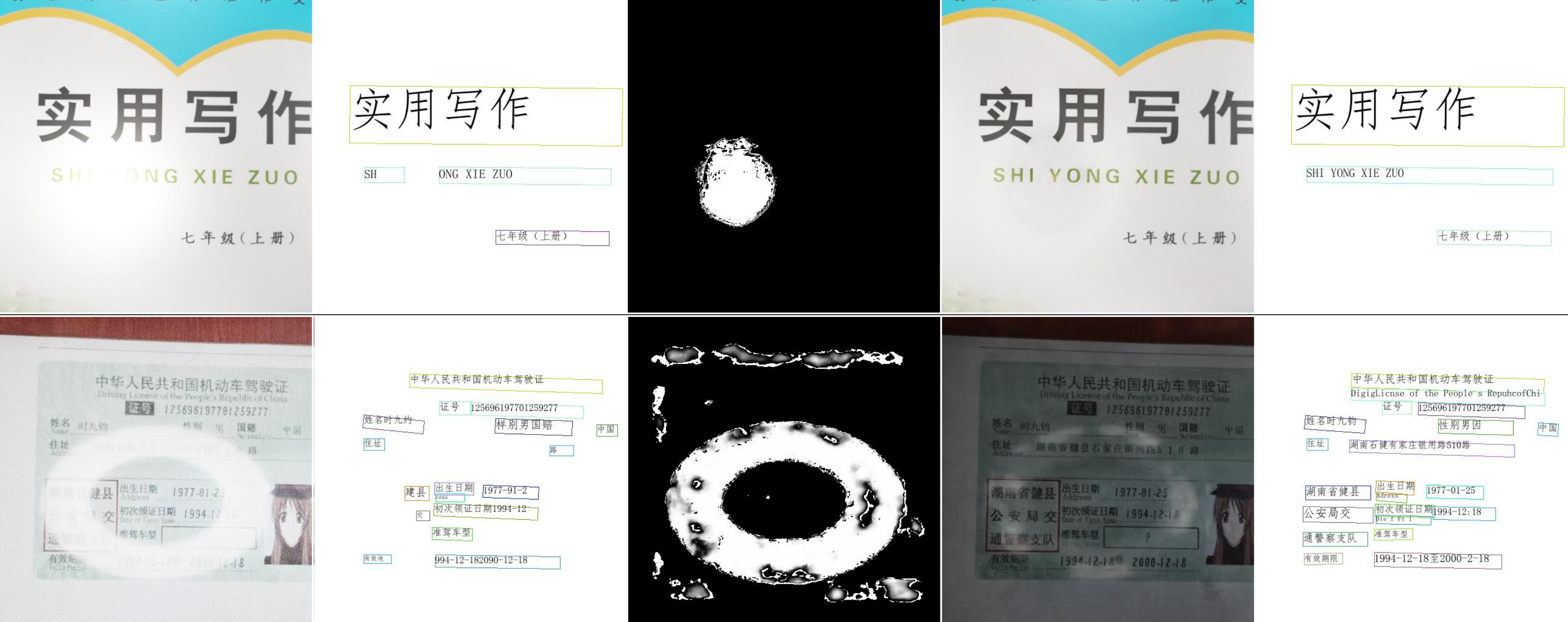}
    \put(-330,-10){{Highlight Image}}
    \put(-255,-10){{OCR Result}}
    \put(-175,-10){{Mask}}
    \put(-110,-10){{Ours}}
    \put(-55,-10){{OCR Result}}
\caption{Selected single image specular highlight removal results of our method. ``Mask'' and ``Ours'' separately are the outputs of our first-stage and second-stage networks.}
\label{fig:examples}
\vskip -0.4cm
\end{figure}
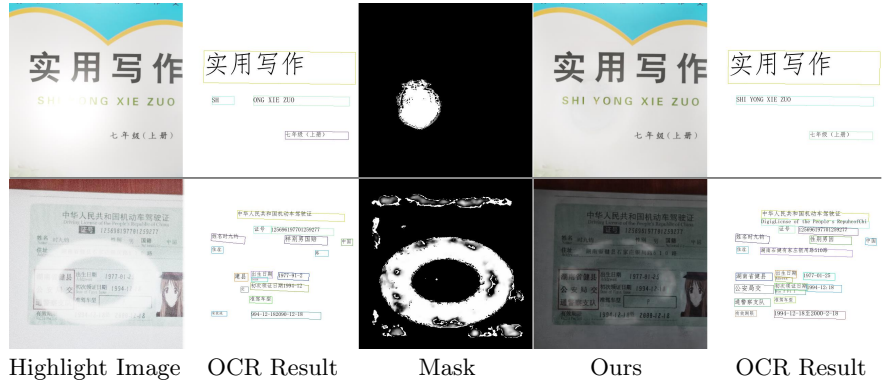

Specular highlights often exist in real-world images due to the material property of objects and the capturing environments. It is always desired to reduce or eliminate these specular highlights to improve the visual quality and to facilitate the vision and graphics tasks, such as  stereo matching~\cite{guo2019deep,khanian2017photometric}, text recognition~\cite{long2021scene}, image segmentation~\cite{arbelaez2011contour,fleyeh2006shadow} and photo-consistency~\cite{wang2015occlusion,wang2019image}. See Fig.~\ref{fig:examples} for examples, the performance of the end-to-end text detection and recognition drops due to the existence of highlight in the images, while our method is designed to detect and remove the highlight so as to improve the subsequent OCR performance.

In the last decades, many approaches have been proposed to address this challenging specular highlight removal problem. These existing works can be roughly classified into three categories: dichromatic reflection model-based methods~\cite{tan2004separating,yang2010real,ren2017specular,fu2019specular,son2020toward}, inpainting-based methods~\cite{tan2003highlight,Ortiz2005a,Arnold2010automatic,Meslouhi2011automatic}, and deep learning-based methods~\cite{funke2018generative,lin2019deep,muhammad2020spec}. The dichromatic reflection model~\cite{shafer1985using} linearly combines the diffuse and specular reflections, and subsequently many methods are proposed based on this model. These methods usually require some simplifying assumptions. In addition, they often need to carry out the pre-processing operations, \eg, segmentation, when encountering images with diverse colors and complex textures, which results in low efficiency and weak practicability. Inpainting-based methods mainly recover the original image contents behind the highlight borrowing the techniques from the image inpainting community. This kind of methods have limited performance for the large highlight contamination. Considering the complexity of single image specular highlight removal, some recent works~\cite{funke2018generative,lin2019deep,muhammad2020spec} are proposed based on the deep neural networks, \eg, convolutional neural network (CNN) and generative adversarial network (GAN). With the aid of the powerful learning capacity of deep models, these deep learning-based methods usually have better performance compared with traditional optimization-based methods. However, these deep learning-based methods require the large-scale training data, especially paired real-world images with necessary annotations, which are time-consuming even difficult to collect. 

Existing specular highlight removal methods mainly process the medical images, natural images, and specific-object images, however, there is no work to focus on the text images. On the other hand, for the end-to-end text detection and recognition, many approaches are proposed to handle texts with arbitrary shapes and various orientations. To our knowledge, the case of text images with specular highlight contamination is rarely studied. Therefore, in this paper, we conduct an extensive study on the text-aware single image specular highlight removal problem including dataset collection, network architecture, training losses, and evaluation metrics. 
The main contributions of our work are as follows: 
\begin{itemize}
    \item We first raise and study in the literature text-aware single image specular highlight removal problem. To study this challenging problem, we collect three high-quality datasets with fine-grained annotations.
    \item We propose a novel two-stage framework of highlight regions detection and removal implemented with two sub-networks. The highlight detection network provides the useful location information to facilitate the subsequent highlight removal network. For the training objectives, we jointly exploit detection loss, reconstruction loss, GAN loss, and text-related loss to achieve the good performance.
    \item For the result comparison, we suggest a comprehensive measurement set, which contains the end-to-end text detection and recognition performance and auxiliary visual quality evaluation.
\end{itemize}

\section{Related Work} 

\subsection{Dichromatic Reflection Model-Based Methods}
The dichromatic reflection model~\cite{shafer1985using} assumes that the image intensity can be represented by a linear combination of diffuse and specular reflections. This model have been widely used for specular highlight removal. Based on the distribution of diffuse and specular points in the maximum chromaticity-intensity space, Tan \etal~\cite{tan2004separating} separated the reflection components by identifying the diffuse maximum chromaticity and then applying a specular-to-diffuse mechanism. Inspired by the observation that diffuse maximum chromaticity in a local patch of color images changes smoothly, Yang \etal~\cite{yang2010real} enhanced the real-time performance and robustness of the chromaticity estimation by applying the bilateral filtering. To exploit the global information of color images for specular reflection separation, Ren \etal~\cite{ren2017specular} proposed a global color-lines constraint based on the dichromatic reflection model. Fu \etal~\cite{fu2019specular} reformulated estimating the diffuse and specular images as an energy minimization with sparse constraints, which can be approximately solved. Recently, Son \etal~\cite{son2020toward} proposed a convex optimization framework to effectively remove the sepcular highlight from chromatic and achromatic regions of natural images. These dichromatic reflection model-based approaches often have limited performance for processing the images with diverse colors and complex textures.

\subsection{Inpainting-Based Methods}
Inpainting is to complete the missing regions of images by propagating information from the known regions, and this technique can be used to restore damaged paintings or remove specific objects~\cite{bertalmio2000image}. Tan \etal~\cite{tan2003highlight} first proposed an inpainting-based method for highlight removal by incorporating the illumination-based constraints. Ortiz and Torres~\cite{Ortiz2005a} designed a connected vectorial filter integrating into the inpainting process to eliminate the specular reflectance. Park and Lee~\cite{park2007inpainting} introduced a highlight inpainting method based on the color line projection, however, this method needs two images taken with different exposure times. Inpainting-based highlight removal methods were also proposed to 
handle the medical images, such as endoscopic images~\cite{Arnold2010automatic} and colposcopic images~\cite{Meslouhi2011automatic}. However, these inpainting-based methods are only effective for images with small areas of highlight contamination. 

\subsection{Deep Learning-Based Methods}
Different from the aforementioned two kinds of methods, the deep learning-based methods do not require the specular highlight model assumption, and thus have the potential to handle various scenarios. Lee \etal~\cite{lee2010removal} proposed a perceptron artificial neural network to detect the specular reflections of tooth images and then applied the smoothing spatial filter to recursively correct the specular reflections. Due to the lack of paired training data, Funke \etal~\cite{funke2018generative} adopted the cycle GAN framework~\cite{zhu2017unpair} and introduced a self-regularization loss aiming to reduce image modification in non-specular regions. Similarly, Lin \etal~\cite{lin2019deep} also adopted a GAN framework and proposed a multi-class discriminator, where classifying the generated diffuse images from real ones and original input images as well. Muhammad \etal~\cite{muhammad2020spec} proposed two deep models (Spec-Net and Spec-CGAN) for specularity removal from faces. The former takes the intensity channel as input while the latter takes the RGB image as input. These methods mainly proposed for the medical images, specific-object images or facial images, whereas our work pays attention to the text images.

\begin{figure}[t]
\centering
\includegraphics[width=0.80\linewidth]{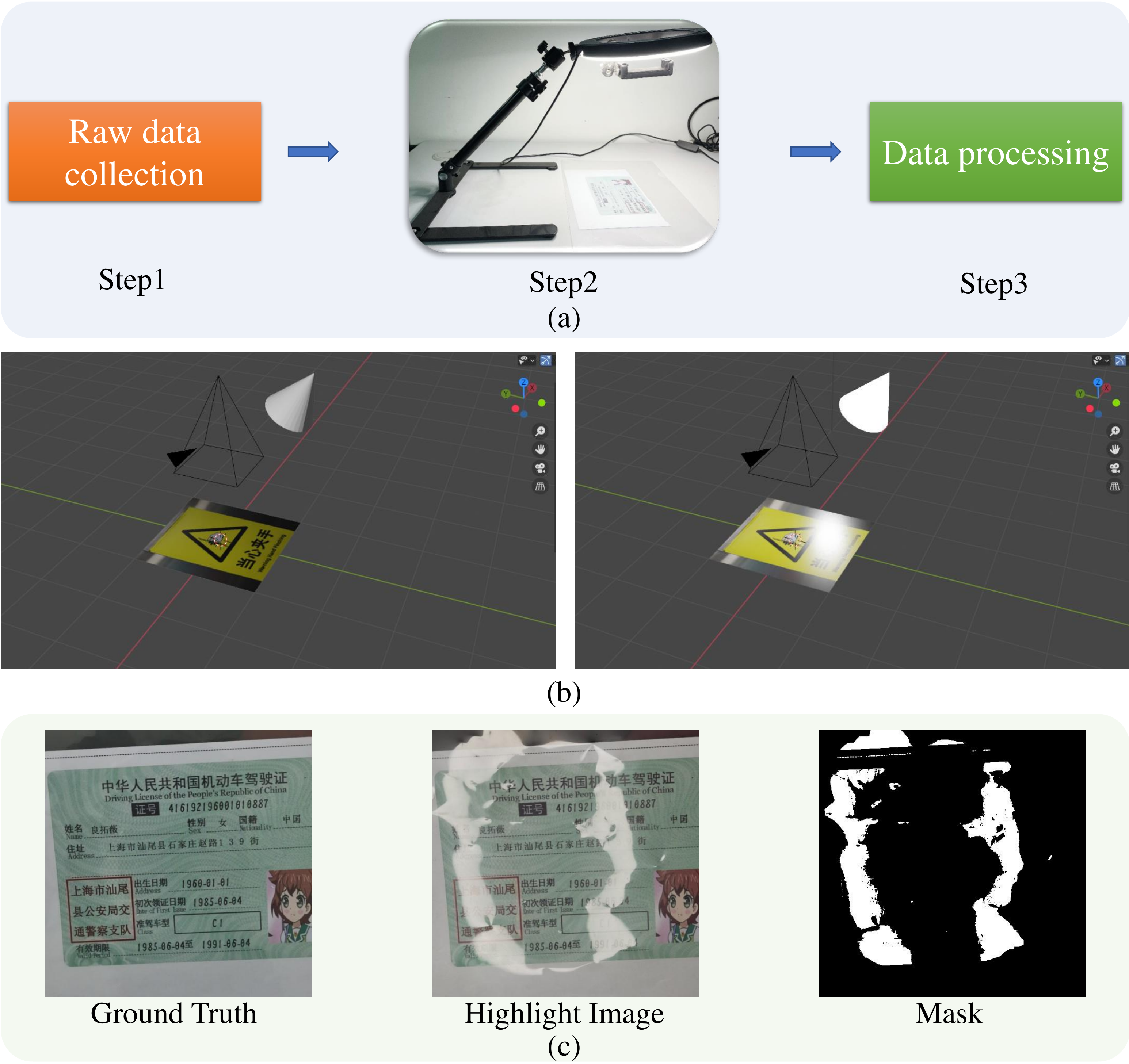}
\caption{The collection pipelines of real dataset (a) and synthetic dataset (b), and an example of paired data sample (c).}
\label{fig:Dataset_process}
\vskip -0.45cm
\end{figure}

\section{Datasets}
In the literature, there is no publicly available dataset for studying the text-aware single image specular highlight removal problem. Therefore, in this work, we collect three high-quality datasets including a real dataset and two synthetic datasets. The pipelines of datasets collection and an example of paired data sample are shown in Fig.~\ref{fig:Dataset_process}.

\subsection{Real Dataset}
\label{subsec:real_data}

Fig.~\ref{fig:Dataset_process}(a) illustrates the pipeline of real dataset collection. For the real dataset, we collect 2,025 image pairs: image with text-aware specular highlight, the corresponding highlight-free image and binary mask image indicating the location of highlight. The image contents include ID cards and driver's licenses, which contain a lot of text information. We first put the transparent plastic film on the picture and turn on the light. Then, the camera shoots to obtain a highlight image. Correspondingly, we obtain a highlight-free image by turning off the light. The shapes and intensities of the highlights are various by adjusting the location of the plastic film. Binary mask image is achieved from the image with specular highlight and highlight-free image through difference and multiple threshold screening. We randomly split this dataset (named RD) into a training set (1,800 images) and a test set (225 images).

\subsection{Synthetic Datasets}
To further enrich the diversity of our dataset, we construct two sets of synthetic images using the 3D computer graphics software Blender. Fig.~\ref{fig:Dataset_process}(b) shows the pipeline of synthetic dataset collection. We first collect 3,679 images with texts from supermarkets and streets, and 2,025 images mentioned in Sec.~\ref{subsec:real_data}. Then, we use the Blender with Cycles engine to automatically generate 27,700 groups of text-aware specular highlight images, the corresponding highlight-free images and highlight mask images. In particular, the highlight shapes include circles, triangles, ellipses, and rings to simulate the lighting conditions in real-world scenes. The material roughness is randomly set in the range [0.1,0.3], and the illumination intensity is randomly chosen from the range [40,70]. To force the specular highlight on the text areas of the image, we provide the Blender with the location information of the text areas obtained via the text detection model CTPN~\cite{tian2016detecting}. 

Because the product or street view category contains less texts per image, while the texts in ID cards and driver's licenses are more dense. Under the same illumination condition, the difficulty of restoring the text information interfered by the specular highlight in these two kinds of images is different. Therefore, we divide the above two types of images into two datasets, namely, SD1 and SD2. SD1 contains 12,000 training sets and 2,000 test sets. SD2 contains 12,000 training sets and 1,700 test sets. Note that, the image contents of RD and SD2 are same.

\section{Proposed Method}
\begin{figure}[t]
\centering
\includegraphics[width=0.95\linewidth]{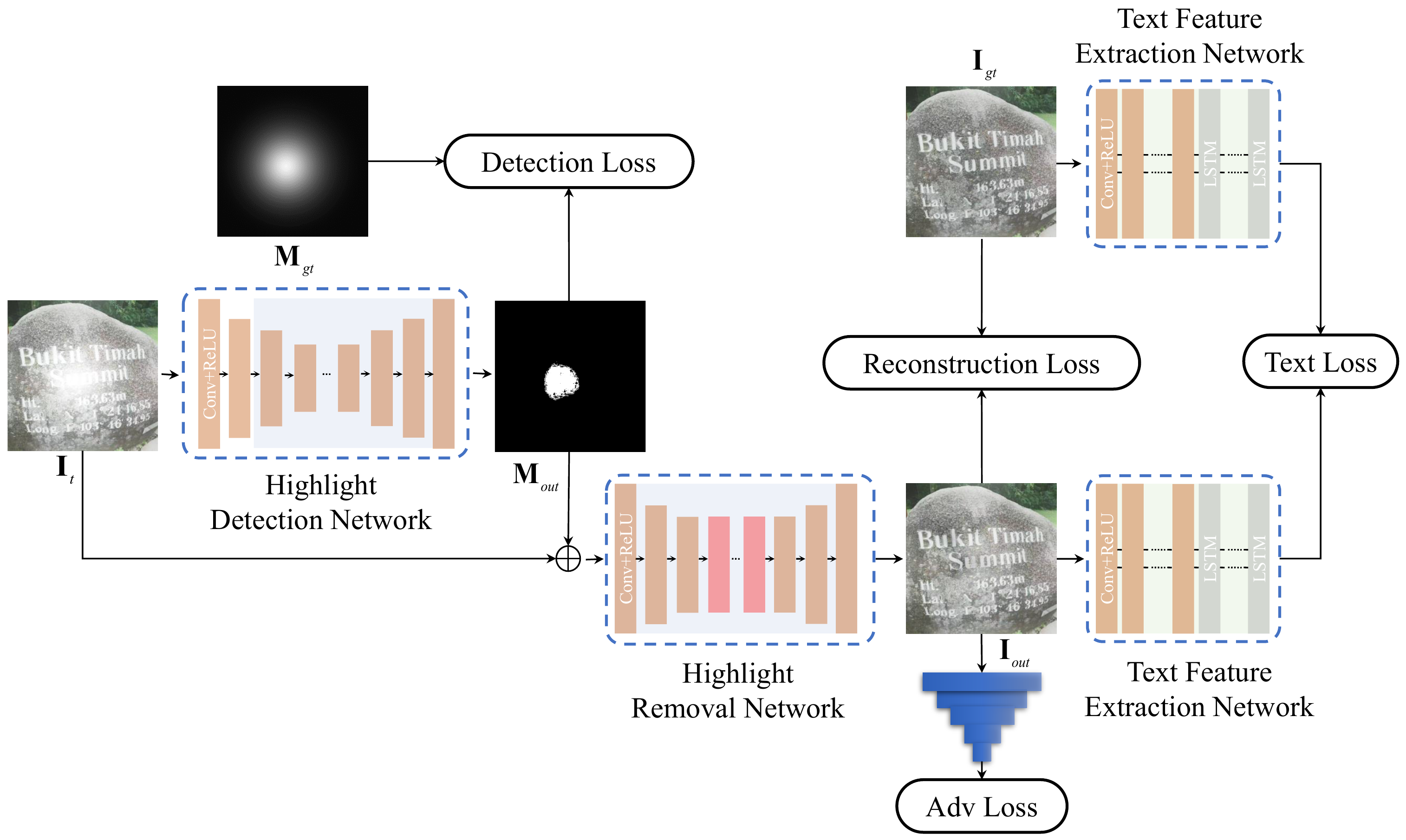}
\caption{The whole structure of our proposed specular highlight removal framework, which consists of a highlight detection network, a highlight removal network, and a patch-based discriminator. Symbol $\bigoplus$ means the channel-wise concatenation.}
\label{fig:framework}
\vskip -0.5cm
\end{figure}

In this work, we propose a two-stage framework to detect and remove the specular highlight from text images. The whole architecture is shown in Fig.~\ref{fig:framework}. In the following, we describe the details of our network architecture and the loss functions.

\subsection{Network Architecture}

\paragraph{\textbf{Highlight Detection Network.}} The highlight detection network $Net_D$ takes the text image $\mathbf{I}_t$ with specular highlight as input and outputs a mask $\mathbf{M}_{out}$ indicating the highlight regions. Each element of $\mathbf{M}_{out}$ is in [0,1], and a larger value stands for a higher probability that the corresponding location of image $\mathbf{I}_t$ is covered by the specular highlight. Due to the same width and height of $\mathbf{I}_t$ and $\mathbf{M}_{out}$, for this network, we adopt a fully convolutional architecture consisting of three downsampling and upsampling layers. Each downsampling layer is followed by two convolutional layers, and each upsampling layer is followed by three convolutional layers.

\paragraph{\textbf{Highlight Removal Network.}}
After achieving the highlight mask $\mathbf{M}_{out}$, the highlight removal network $Net_R$ is then applied to remove the specular highlight and recover the text information. As input, $Net_R$ accepts an input text image $\mathbf{I}_t$ and detected highlight mask $\mathbf{M}_{out}$. The output of our highlight removal network $Net_R$ is a highlight-free image $\mathbf{I}_{out}$. Through introducing $\mathbf{M}_{out}$, network $Net_R$ can pay more attention to the highlight regions and achieve more better removal performance. For the network architecture of $Net_R$, in this work, we adopt an encoder-decoder network with skip connection. This network consists of two downsampling layers, four residual blocks, and two upsampling layers. To further enhance the removal performance, we also apply a patch-based discriminator~\cite{miyato2018spectral}. The discriminator $D$ includes one convolutional layer and five downsampling layers with kernel size of 5 and stride of 2. The spectral normalization is utilized to stabilize the training of the discriminator.

\subsection{Loss Functions}
Next, we illustrate the loss functions used for training our network.

\paragraph{\textbf{Highlight Detection Loss.}} For the objective function of highlight detection network, we use $l_1$ loss, \ie, $\mathcal{L}_{Net_D} = \| \mathbf{M}_{out} - \mathbf{M}_{gt} \|_{1}$, where $\mathbf{M}_{gt}$ is the ground truth of highlight mask.

\paragraph{\textbf{Reconstruction Loss.}}
The reconstruction loss is to add constraints on the pixel and feature space. The pixel-aware loss consists of pixel-wise difference item and total varition (TV) item: $\mathcal{L}_{P} = 5 * \| \mathbf{I}_{out} - \mathbf{I}_{gt} \|_{1} + 0.1 * (\| \mathbf{I}_{out}(i,j) - \mathbf{I}_{gt}(i-1,j) \|_{1} + \| \mathbf{I}_{out}(i,j) - \mathbf{I}_{gt}(i,j-1) \|_{1})$. The feature-aware loss including perceptual loss~\cite{johnson2016perceptual} and style loss~\cite{Gatys2016image}: $\mathcal{L}_{F} = 0.05 * \| \Phi(\mathbf{I}_{out}) - \Phi(\mathbf{I}_{gt}) \|_{1} + 120 * \| \Psi(\mathbf{I}_{out}) - \Psi(\mathbf{I}_{gt}) \|_{1}$, where $\Phi$ is the feature maps of pre-trained VGG-16~\cite{simonyan2014very} and $\Psi(\cdot) = \Phi(\cdot)\Phi(\cdot)^{T}$ is the Gram matrix~\cite{Gatys2016image}. The feature-aware loss improves the visual quality of results.

\paragraph{\textbf{GAN Loss.}}
In the highlight removal network, we use a patch-based discriminator $D$ to enhance the visual realism of results. For the GAN loss, we adopt the hinge loss. Therefore, the adversarial loss for $Net_R$ is $\mathcal{L}_G = - \mathbb{E} [ D(\mathbf{I}_{out}) ]$. The loss used for training the discriminator $D$ is formulated as $\mathcal{L}_D = \mathbb{E} [max(0, 1 - D(\mathbf{I}_{gt}))] + \mathbb{E}[ max(0, 1 + D(\mathbf{I}_{out}))]$.

\paragraph{\textbf{Text-Related Loss.}}
In this work, our specular highlight removal is text-aware. This means that the highlight removal network $Net_R$ needs to pay more attention to recover the texts hidden behind the highlights. To do this, we apply the pre-trained text detection and recognition models to provide the supervision on the text recovering. More specifically, we add the consistent constraints on the feature maps of $\mathbf{I}_{out}$ and $\mathbf{I}_{gt}$ extracted from above two pre-trained models, and the text-related loss is formulated as $\mathcal{L}_{T} = \sum_{c=1}^{3}  \|\phi_{c}(\mathbf{I}_{out})-\phi_{c}(\mathbf{I}_{gt})\|_1 + \|\phi_{d}(\mathbf{I}_{out})-\phi_{d}(\mathbf{I}_{gt})\|_1$, where $\phi_{c}$ stands for the $c$-th layer feature map from the pre-trained CTPN model~\cite{ChineseOCR} and $\phi_{d}$ denotes the $d$-th layer feature map from the pre-trained DenseNet~\cite{ChineseOCR}.

To this end, the total objective function of $Net_D$ and $Net_R$ is $\mathcal{L} = \lambda_{Net_D} \mathcal{L}_{Net_D} + \mathcal{L}_{P} + \mathcal{L}_{F} + \lambda_{G} \mathcal{L}_{G} + \mathcal{L}_{T}$. In all experiments, we set $\lambda_{Net_D} = 10$ and $\lambda_{G} = 0.01$.

\section{Experiments}

\subsection{Implementation Settings}
Our network is implemented with TensorFlow 1.15. As GPU we use a TITAN RTX from NVIDIA\textsuperscript{\textregistered}. The Adam optimizer~\cite{kingma2015adam} with a batch size of 4 is used to train our network, where $\beta_1 = 0.5$ and $\beta_2 = 0.9$. The learning rate is initialized as 0.0001. In our experiments, all the images are of size of $512 \times 512$. Note that, the text recognition model used for result evaluation is different from the model used in text-related loss.

\subsection{Qualitative Evaluation}
\begin{figure}[!ht]
\centering
\includegraphics[width=0.94\linewidth]{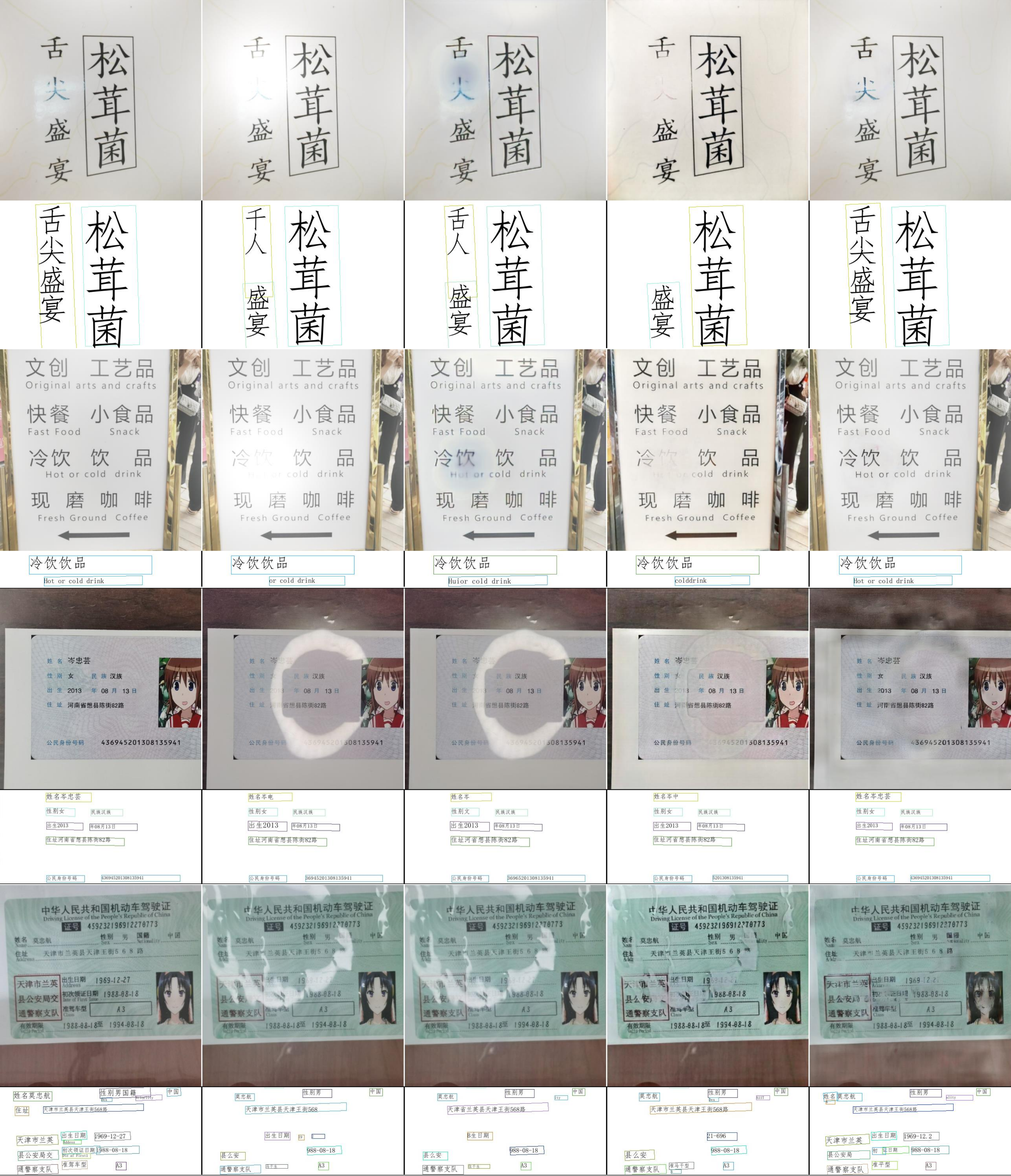}
    \put(-322,-10){{Ground Truth}}
    \put(-260,-10){{Highlight Image}}
    \put(-175,-10){{Multi}}
    \put(-110,-10){{SPEC}}
    \put(-45,-10){{Ours}}
\caption{Qualitative comparisons of our method with Multi~\cite{lin2019deep} and SPEC~\cite{muhammad2020spec}.} 
\label{fig:Qualitative_comparisons}
\vskip -0.4cm
\end{figure}
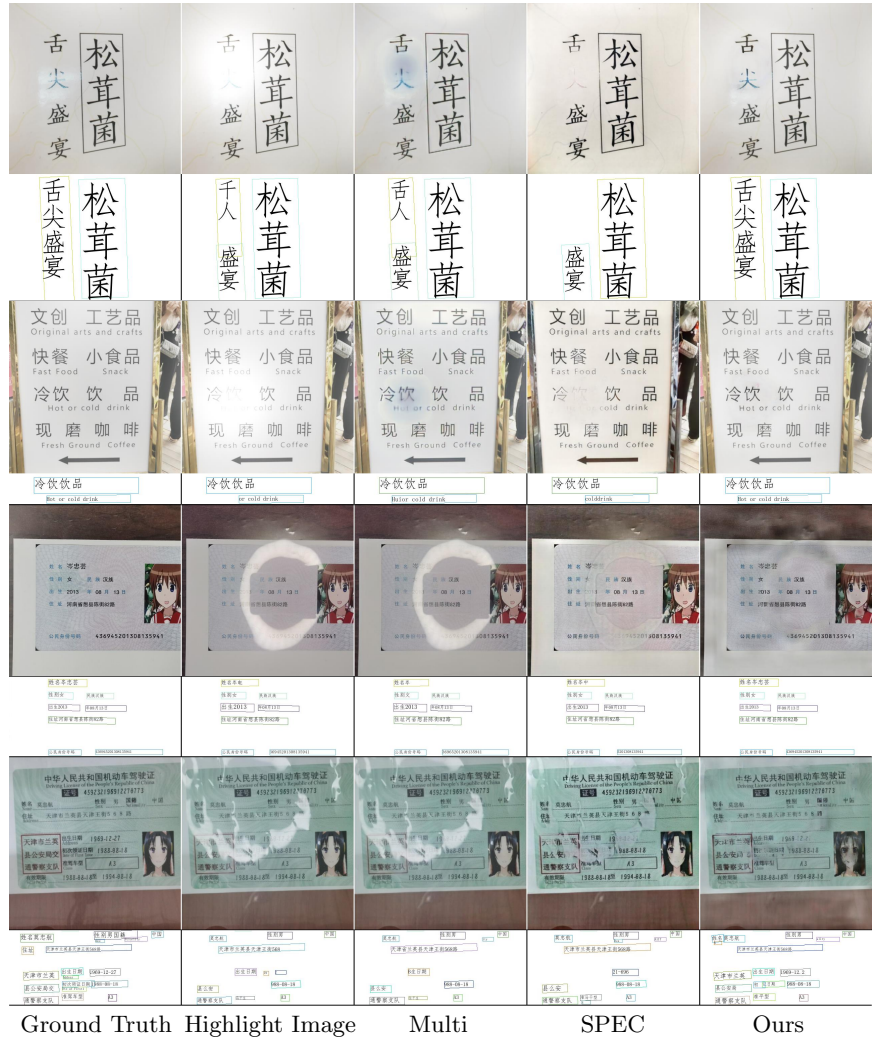

We qualitatively compare our method with two recent advanced specular highlight removal methods: Multi~\cite{lin2019deep} and SPEC (Spec-CGAN~\cite{muhammad2020spec}) on our collected three datasets. The results are shown in Fig.~\ref{fig:Qualitative_comparisons}. Among these three methods, our method can better remove the highlight and achieve the superior end-to-end text detection and recognition performance. For example, our method successfully recovers the name, address, and id number in the third row. Multi has apparent highlight remnants in the third and fourth rows due to its blind removal property, whereas our method can better perceive the highlight regions. Compared with Multi, the results of SPEC have less highlights, however, the capability of recovering the texts is limited for the cycleGAN framework as SPEC followed.

\subsection{Quantitative Evaluation}
In addition, we quantitatively compare the above three methods in terms of the end-to-end text detection and recognition performance and visual quality. For the end-to-end text detection and recognition evaluation, we adopt the common metrics~\cite{lucas2003icdar}: recall, precision, and f-measure. We choose the current advanced text detection and recognition algorithm Paddle OCR~\cite{PaddleOCR} to calculate these three metrics. For visual quality evaluation, we utilize the PSNR and SSIM. 

Table~\ref{tab:comp} reports the numerical results of the three methods on our three datasets. Due to the same image contents of RD and SD2, for real dataset (RD), we fine-tune the model trained on SD2 using the training set of RD for all three methods. From Table~\ref{tab:comp}, we can find that our method achieves the best performance for end-to-end text detection and recognition (see 3-5 columns). Take the recall as an example, our method can improve the end-to-end text detection and recognition performance by 6.89\% (SD1), 3.07\% (SD2), and 13.65\% (RD), respectively. This improvement indicates that our method can better recover the original texts hidden behind the specular highlight. In addition, the end-to-end detection and recognition performance of Multi and SPEC sometimes is lower than that of Light Image. The reason is that these two methods remove the highlight and texts as well. For PSNR and SSIM, SPEC is worst, while our method and Multi are competitive for synthetic datasets, and our method is better than Multi for real datasets. PSNR and SSIM of our method sometimes are lower than that of Multi, however, these two metrics are not exactly the same as the visual quality that the human eyes perceive. More importantly, we focus on the end-to-end text detection and recognition performance after highlight removal, and the visual quality is an auxiliary aspect.

\begin{table*}[t]
\caption{Quantitative comparison of our method and two recent state-of-the-art methods: Multi~\cite{lin2019deep} and SPEC~\cite{muhammad2020spec}. All three methods are trained and tested on our collected three datasets separately. Recall, precision, f-measure, and SSIM are in $\%$.}
\label{tab:comp}
\centering
\footnotesize
\setlength{\tabcolsep}{4pt}
\begin{tabular}{cccccccccccccc}
\hline
Datasets & Methods & Recall ↑ & Precision ↑ & F-measure ↑ & PSNR ↑ & SSIM ↑ & \\ \hline
\multirow{4}{*}{SD1}
     & Light Image & 85.03 & 94.70 & 88.70 & 17.58 & 82.37  \\ 
  & Multi(2019) & 86.28 & 94.76 & 89.30 & \textbf{26.29} & \textbf{89.86}   \\ 
   & SPEC(2020) & 82.39 & 93.12 & 86.31 & 15.61 & 68.82  \\ 
 & \textbf{Ours} & \textbf{91.92} & \textbf{96.32} & \textbf{93.57} & 22.65 & 88.33   \\ \hline \hline
 \multirow{4}{*}{SD2}
 & Light Image & 80.50 & \textbf{95.89} & 87.10 & 11.79 & 66.42  \\ 
  & Multi(2019) & 79.21 & 93.82 & 84.88 & 28.99 & \textbf{91.81}  \\ 
    & SPEC(2020) & 78.87 & 95.10 & 85.55 & 9.66 & 53.95  \\ 
 & \textbf{Ours} &\textbf{83.57} & 95.00 & \textbf{88.42} & \textbf{29.21} & 90.67  \\ \hline \hline
  \multirow{4}{*}{RD}
 & Light Image & 64.85 & 90.60 & 73.49 & 17.05 & 65.04  \\ 
  & Multi(2019) & 61.58 & 87.63 & 70.72 & 17.17 & 64.23   \\ 
    & SPEC(2020) & 70.59 & \textbf{91.62} & 78.38 & 14.82 & 52.49  \\ 
 & \textbf{Ours} & \textbf{78.50} & 91.34 & \textbf{83.34} & \textbf{21.62} & \textbf{77.19}   \\ \hline
\end{tabular}
\end{table*}

\subsection{Ablation Study}
To verify the effectiveness of the text-related loss, we perform the ablation experiments and report the corresponding results in Table~\ref{tab:Ablation_loss}. We observe that the end-to-end text detection and recognition performance of our method with text-related loss is consistently improved for three datasets. This indicates that the text-related loss can enforce the highlight removal network to conduct the text-aware restoration. In addition, we can find that the end-to-end text detection and recognition performance of our method is already better than that of Multi and SPEC (comparing the first row of each dataset in Table~\ref{tab:Ablation_loss} with the corresponding rows in Table~\ref{tab:comp}), even though there is no text-related loss. 

\begin{table*}[t]
\caption{Performance of our method without and with text-related loss on our collected three datasets.} 
\label{tab:Ablation_loss}
\footnotesize
\centering
\setlength{\tabcolsep}{4pt}
\begin{tabular}{cccccccccccccc}
\hline
Datasets & Methods & Recall ↑ & Precision ↑ & F-measure ↑ & PSNR ↑ & SSIM ↑ & \\ \hline
\multirow{2}{*}{SD1}
    & w/o text loss & 91.43 & 94.12 & 92.75 & 21.88 & 87.19  \\ 
 & \textbf{Ours} & \textbf{91.92} & \textbf{96.32} & \textbf{93.57} & \textbf{22.65} & \textbf{88.33}   \\ \hline \hline
 \multirow{2}{*}{SD2}
    & w/o text loss & 82.69 & 93.48 & 87.76 & 28.12 & 89.93  \\ 
 & \textbf{Ours} & \textbf{83.57} & \textbf{95.00} & \textbf{88.42} & \textbf{29.21} & \textbf{90.67}  \\ \hline \hline
 \multirow{2}{*}{RD}
   & w/o text loss & 77.04 & 89.66 & 82.87 & 21.38 & 76.11  \\ 
 & \textbf{Ours} & \textbf{78.50} & \textbf{91.34} & \textbf{83.34} & \textbf{21.62} & \textbf{77.19}   \\ \hline
\end{tabular}
\vskip -4mm
\end{table*}

\section{Conclusion and Future Work}
In this work, we studied and solved the challenging specular highlight removal problem of single text image. To facilitate this study, we collected three high-quality datasets with fine-grained annotations. We proposed a two-stage framework including a highlight detection network and a highlight removal network. The output of highlight detection network is used as an auxiliary information, which guides the highlight removal network to pay more attention to the highlight regions. In addition, text-related loss was introduced to improve the recovering of texts. Our source code and datasets are available at\\ \re{\url{https://github.com/weizequan/TASHR}}.

In the future, we would like to construct lager and richer dataset to promote the development of related research. We would also like to design more effective networks and loss functions. Furthermore, an exciting research problem is to suggest more complete and exact visual quality measurements. 

\section*{Acknowledgements}
This work was supported by the National Key R\&D Program of China (2019YFB2204104) and the National Natural Science Foundation of China (61772523).

%
%
%
%
{

\bibliographystyle{splncs04}

\bibliography{refer}

\begin{thebibliography}{10}
\providecommand{\url}[1]{\texttt{#1}}
\providecommand{\urlprefix}{URL }
\providecommand{\doi}[1]{https://doi.org/#1}

\bibitem{ChineseOCR}
Chineseocr: Ctpn plus densenet plus ctc based chinese ocr,
  {https://github.com/YCG09/chinese\_ocr}. Accessed 30 April 2021.

\bibitem{PaddleOCR}
Paddleocr: Awesome multilingual ocr toolkits based on paddlepaddle,
  {https://github.com/PaddlePaddle/PaddleOCR}. Accessed 30 April 2021.

\bibitem{arbelaez2011contour}
Arbel\'aez, P., Maire, M., Fowlkes, C., Malik, J.: Contour detection and
  hierarchical image segmentation. IEEE Transactions on Pattern Analysis and
  Machine Intelligence  \textbf{33}(5),  898--916 (2011)

\bibitem{Arnold2010automatic}
Arnold, M., Ghosh, A., Ameling, S., Lacey, G.: Automatic segmentation and
  inpainting of specular highlights for endoscopic imaging. Journal on Image
  and Video Processing  \textbf{2010} (2010)

\bibitem{bertalmio2000image}
Bertalmio, M., Sapiro, G., Caselles, V., Ballester, C.: Image inpainting. In:
  ACM {SIGGRAPH}. pp. 417--424 (2000)

\bibitem{fleyeh2006shadow}
Fleyeh, H.: Shadow and highlight invariant colour segmentation algorithm for
  traffic signs. In: IEEE Conference on Cybernetics and Intelligent Systems
  (2006)

\bibitem{fu2019specular}
Fu, G., Zhang, Q., Song, C., Lin, Q., Xiao, C.: Specular highlight removal for
  real-world images. Computer Graphics Forum  \textbf{38}(7),  253--263 (2019)

\bibitem{funke2018generative}
Funke, I., Bodenstedt, S., Riediger, C., Weitz, J., Speidel, S.: Generative
  adversarial networks for specular highlight removal in endoscopic images. In:
  Medical Imaging 2018: Image-Guided Procedures, Robotic Interventions, and
  Modeling. vol. 10576, pp. 8 -- 16 (2018)

\bibitem{Gatys2016image}
Gatys, L.A., Ecker, A.S., Bethge, M.: Image style transfer using convolutional
  neural networks. In: IEEE Conference on Computer Vision and Pattern
  Recognition. pp. 2414--2423 (2016)

\bibitem{guo2019deep}
Guo, X., Chen, Z., Li, S., Yang, Y., Yu, J.: Deep eyes: Binocular
  depth-from-focus on focal stack pairs. In: Chinese Conference on Pattern
  Recognition and Computer Vision. pp. 353--365 (2019)

\bibitem{johnson2016perceptual}
Johnson, J., Alahi, A., Fei-Fei, L.: Perceptual losses for real-time style
  transfer and super-resolution. In: Proceedings of the European Conference on
  Computer Vision. pp. 694--711 (2016)

\bibitem{khanian2017photometric}
Khanian, M., Boroujerdi, A.S., Breu{\ss}, M.: Photometric stereo for strong
  specular highlights. arXiv preprint arXiv:1709.01357  (2017)

\bibitem{kingma2015adam}
Kingma, D.P., Ba, J.: Adam: A method for stochastic optimization. In:
  International Conference on Learning Representations (2015)

\bibitem{lee2010removal}
Lee, S.T., Yoon, T.H., Kim, K.S., Kim, K.D., Park, W.: Removal of specular
  reflections in tooth color image by perceptron neural nets. In: International
  Conference on Signal Processing Systems. vol.~1, pp. V1--285--V1--289 (2010)

\bibitem{lin2019deep}
Lin, J., El~Amine~Seddik, M., Tamaazousti, M., Tamaazousti, Y., Bartoli, A.:
  Deep multi-class adversarial specularity removal. In: Image Analysis. pp.
  3--15 (2019)

\bibitem{long2021scene}
Long, S., He, X., Yao, C.: Scene text detection and recognition: The deep
  learning era. International Journal of Computer Vision  \textbf{129}(1),
  161--184 (2021)

\bibitem{lucas2003icdar}
Lucas, S., Panaretos, A., Sosa, L., Tang, A., Wong, S., Young, R.: Icdar 2003
  robust reading competitions. In: International Conference on Document
  Analysis and Recognition. pp. 682--687 (2003)

\bibitem{Meslouhi2011automatic}
Meslouhi, O.E., Kardouchi, M., Allali, H., Gadi, T., Benkaddour, Y.A.:
  Automatic detection and inpainting of specular reflections for colposcopic
  images. Central European Journal of Computer Science  \textbf{1} (2011)

\bibitem{miyato2018spectral}
Miyato, T., Kataoka, T., Koyama, M., Yoshida, Y.: Spectral normalization for
  generative adversarial networks. In: International Conference on Learning
  Representations (2018)

\bibitem{muhammad2020spec}
Muhammad, S., Dailey, M.N., Farooq, M., Majeed, M.F., Ekpanyapong, M.: Spec-net
  and spec-cgan: Deep learning models for specularity removal from faces. Image
  and Vision Computing  \textbf{93},  103823 (2020)

\bibitem{Ortiz2005a}
Ortiz, F., Torres, F.: A new inpainting method for highlights elimination by
  colour morphology. In: International Conference on Pattern Recognition and
  Image Analysis. pp. 368--376 (2005)

\bibitem{park2007inpainting}
Park, J.W., Lee, K.H.: Inpainting highlights using color line projection. IEICE
  Transactions on Information and Systems  \textbf{90}(1),  250--257 (2007)

\bibitem{ren2017specular}
Ren, W., Tian, J., Tang, Y.: Specular reflection separation with color-lines
  constraint. IEEE Transactions on Image Processing  \textbf{26}(5),
  2327--2337 (2017)

\bibitem{shafer1985using}
Shafer, S.A.: Using color to separate reflection components. Color Research \&
  Application  \textbf{10}(4),  210--218 (1985)

\bibitem{simonyan2014very}
Simonyan, K., Zisserman, A.: Very deep convolutional networks for large-scale
  image recognition. In: International Conference on Learning Representations
  (2015)

\bibitem{son2020toward}
Son, M., Lee, Y., Chang, H.S.: Toward specular removal from natural images
  based on statistical reflection models. IEEE Transactions on Image Processing
   (2020)

\bibitem{tan2003highlight}
Tan, P., Lin, S., Quan, L., Shum, H.Y.: Highlight removal by
  illumination-constrained inpainting. In: IEEE International Conference on
  Computer Vision. pp. 164--169 (2003)

\bibitem{tan2004separating}
Tan, R.T., Nishino, K., Ikeuchi, K.: Separating reflection components based on
  chromaticity and noise analysis. IEEE Transactions on Pattern Analysis and
  Machine Intelligence  \textbf{26}(10),  1373--1379 (2004)

\bibitem{tian2016detecting}
Tian, Z., Huang, W., He, T., He, P., Qiao, Y.: Detecting text in natural image
  with connectionist text proposal network. In: European Conference on Computer
  Vision. pp. 56--72 (2016)

\bibitem{wang2015occlusion}
Wang, T.C., Efros, A.A., Ramamoorthi, R.: Occlusion-aware depth estimation
  using light-field cameras. In: IEEE International Conference on Computer
  Vision. pp. 3487--3495 (2015)

\bibitem{wang2019image}
Wang, W., Deng, R., Li, L., Xu, X.: Image aesthetic assessment based on
  perception consistency. In: Chinese Conference on Pattern Recognition and
  Computer Vision. pp. 303--315 (2019)

\bibitem{yang2010real}
Yang, Q., Wang, S., Ahuja, N.: Real-time specular highlight removal using
  bilateral filtering. In: European Conference on Computer Vision. pp. 87--100
  (2010)

\bibitem{zhu2017unpair}
Zhu, J.Y., Park, T., Isola, P., Efros, A.A.: Unpaired image-to-image
  translation using cycle-consistent adversarial networks. In: IEEE
  International Conference on Computer Vision. pp. 2242--2251 (2017)

\end{thebibliography}
}

\end{document}